\documentclass[a4paper,conference]{IEEEtran}
\usepackage[noadjust]{cite}
\usepackage{hyperref}
\usepackage[pdftex]{graphicx}
\graphicspath{{graphics/}}
\usepackage{amsmath}
\usepackage{array}
\usepackage{lipsum}
\usepackage[caption=false,font=footnotesize]{subfig}
\usepackage{multirow}
\usepackage{tikz}
\usepackage{bm}
\usepackage{mathtools}
\usepackage{amsfonts}
\usepackage{makecell}
\newcommand{\uf}[1]{\underline{#1}}

\DeclareMathOperator*{\mean}{mean}

\newcommand{\roverline}[1]{\displaystyle\mathpalette\doroverline{#1}}
\newcommand{\doroverline}[2]{\overline{#1#2}}

\definecolor{airforceblue}{rgb}{0.36, 0.54, 0.66}

\definecolor{orange}{rgb}{0.60, 0.30, 0.0}

\usepackage[normalem]{ulem}

\begin{document}
\title{MD-Net:\hspace{1.9mm}Multi-Detector\hspace{2.1mm}for\hspace{2.1mm}Local\hspace{2.1mm}Feature\hspace{2.1mm}Extraction}

\author{\IEEEauthorblockN{
Emanuele Santellani \IEEEauthorrefmark{1},
Christian Sormann \IEEEauthorrefmark{1},
Mattia Rossi \IEEEauthorrefmark{2}, 
Andreas Kuhn \IEEEauthorrefmark{2}, 
Friedrich Fraundorfer \IEEEauthorrefmark{1}
}
\IEEEauthorblockA{
\IEEEauthorrefmark{1}
Institute of Computer Graphics and Vision,
Graz University of Technology, Austria \\
Email: \{emanuele.santellani, christian.sormann, fraundorfer\}@icg.tugraz.at}
\IEEEauthorblockA{
\IEEEauthorrefmark{2}
R\&D Center - Stuttgart Laboratory 1,
Sony Europe B.V., Germany. 
Email: \{mattia.rossi, andreas.kuhn\}@sony.com}}

\maketitle

\begin{abstract}

Establishing a sparse set of keypoint correspondences between images is a fundamental task in many computer vision pipelines.
Often, this translates into a computationally expensive nearest neighbor search, where every keypoint descriptor at one image
must be compared with all the descriptors at the others.
In order to lower the computational cost of the matching phase, we propose a deep feature extraction network
capable of detecting a predefined number of complementary sets of keypoints at each image.
Since only the descriptors within the same set need to be compared across the different images, the matching phase
computational complexity decreases with the number of sets.
We train our network to predict the keypoints and compute the corresponding descriptors jointly.
In particular, in order to learn complementary sets of keypoints, we introduce a novel unsupervised loss
which penalizes intersections among the different sets.
Additionally, we propose a novel descriptor-based weighting scheme meant to penalize the detection of keypoints
with non-discriminative descriptors.
With extensive experiments we show that our feature extraction network, trained only on synthetically warped images
and in a fully unsupervised manner, achieves competitive results on 3D reconstruction and re-localization tasks at
a reduced matching complexity.
\end{abstract}

\section{Introduction}
Being able to extract reliable sets of point correspondences between images
is a fundamental requirement for a large variety of computer vision pipelines,
such as Structure from Motion (SfM) \cite{schoenberger2016sfm}, SLAM \cite{orbslam}, Visual Localization \cite{visual_localization}, object detection \cite{object_detection0,object_detection1} and object tracking \cite{object_tracking}.
The problem has been historically divided into two
sequential steps: local feature extraction and pairwise matching.

The feature extraction step starts with the detection of a sparse set of
salient points, referred to as keypoints, in each image.
Objects visible in multiple images should trigger 
the detection of the same set of keypoints, 
in order to permit
the establishment of correspondences between the images.
As a consequence, the
detection process 
is required to be robust to some degree of image alteration,
such as illumination and viewpoint changes or occlusions.
Several algorithms have been proposed during the last decades, classified as either
blob \cite{sift, detector_scaleaffine, detector_scale}, 
corner \cite{harris, fast}
or region detectors \cite{mser}.
The feature extraction step continues with the assignment of a descriptor vector to each detected keypoint, whose purpose is to describe the keypoint neighborhood.
Among the many proposed algorithms for descriptor extraction \cite{sift, orb, surf, brief, brisk}, 
SIFT \cite{sift} and its improved versions \cite{sift_pca, rootsift} 
are the most successful ones and still remain widely used nowadays.
\begingroup
\setlength{\tabcolsep}{0 pt}
\begin{figure}
\centering
\begin{tabular}{cccc}
\includegraphics[width=0.248\linewidth, trim={1.5cm 0.8cm 2.0cm 3.5cm}, clip]{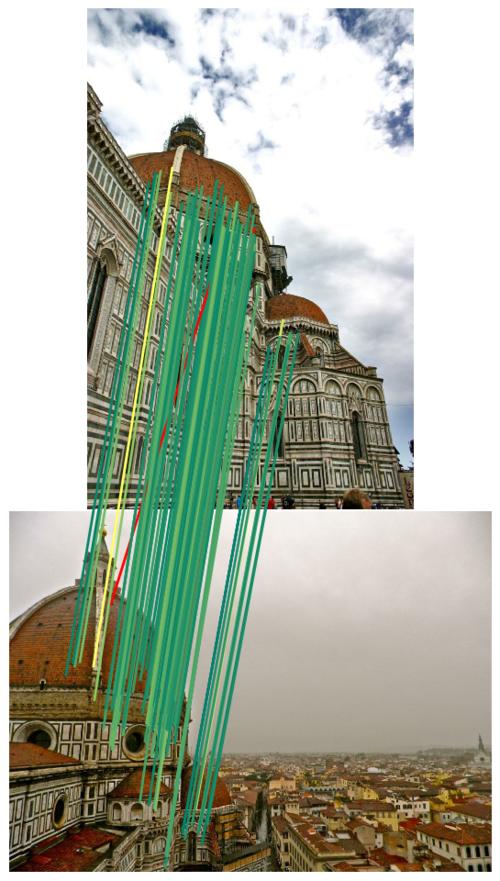} & 
\includegraphics[width=0.248\linewidth, trim={1.5cm 0.8cm 2.0cm 3.5cm}, clip]{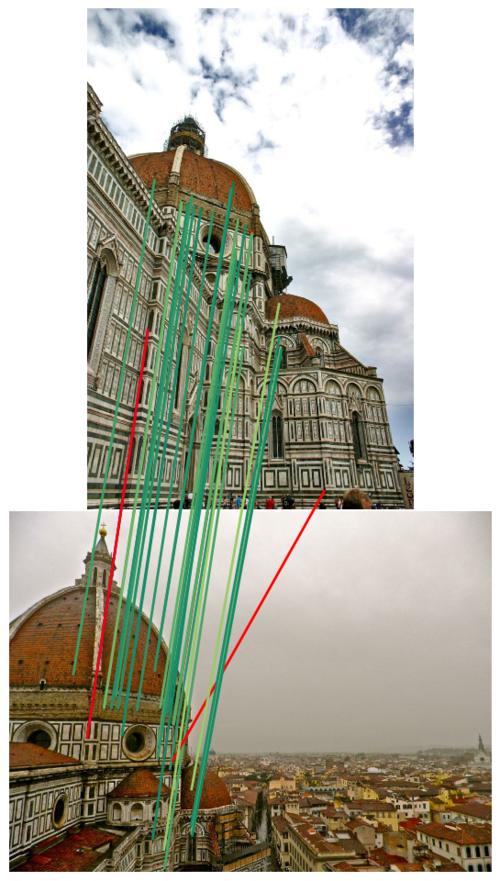} &
\includegraphics[width=0.248\linewidth, trim={1.5cm 0.8cm 2.0cm 3.5cm}, clip]{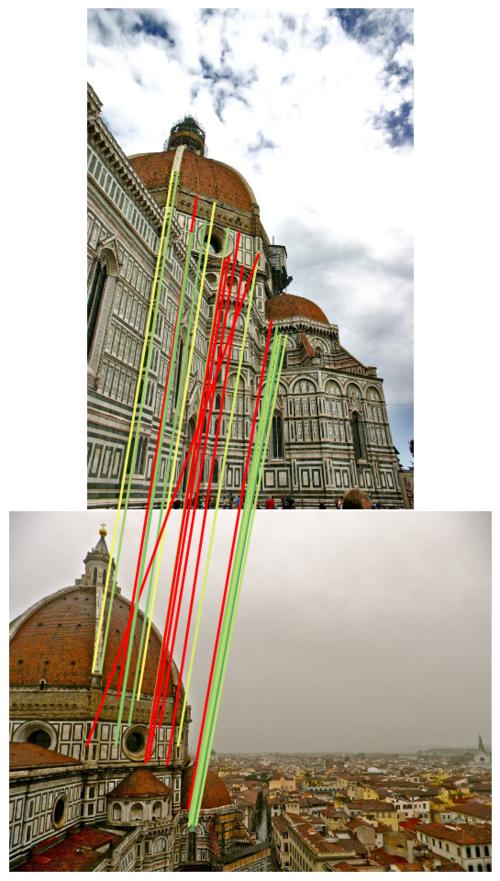} &
\includegraphics[width=0.248\linewidth, trim={1.5cm 0.8cm 2.0cm 3.5cm}, clip]{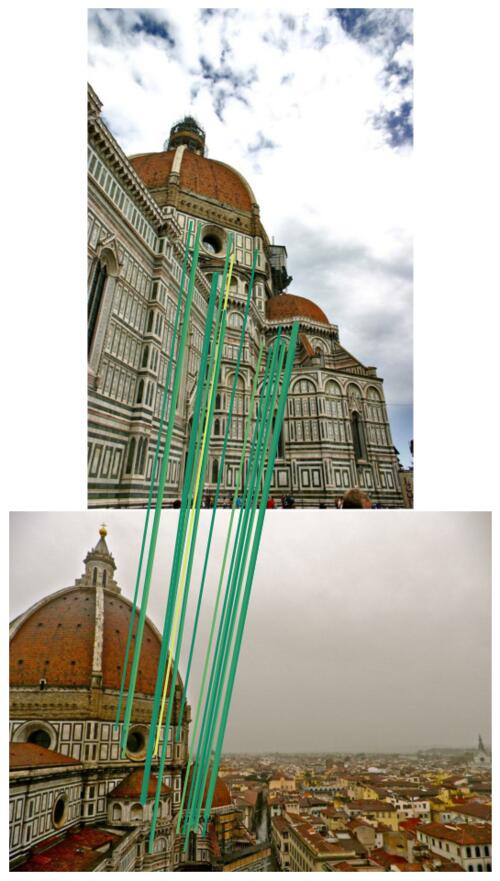} \\
\small{MD-2-Net}\footnotesize{~(ours)} & \small{R2D2 \cite{r2d2}} & \small{ASLFeat \cite{aslfeat}} & \small{Upright-SIFT\,\cite{sift}}
\end{tabular}
\caption{Example pair from the Image Matching Benchmark \cite{imb} stereo task for the scene \textit{Florence Cathedral Side}.
The lines represent the RANSAC inlier matches, color coded according to the pixel error from green,
no error, to red, error above 5px.
}
\label{fig:imb_teaser}
\vspace{-1em}
\end{figure}
\endgroup
\begin{figure*}[!t]
\centering
\vspace*{1ex}
\includegraphics[width=0.8\linewidth, trim={0cm, 3.0cm, 0cm, 1.2cm}, clip]{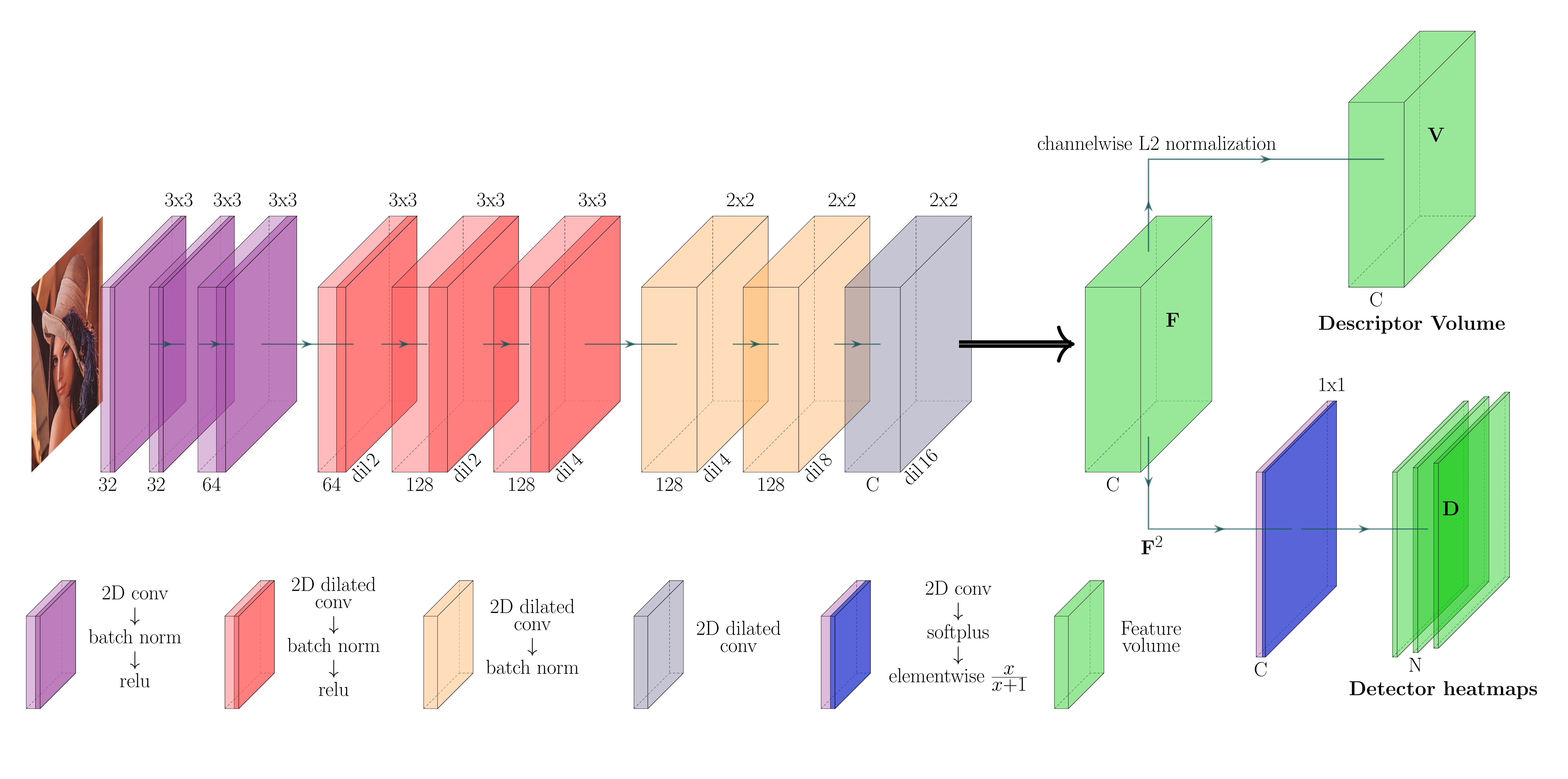}
\caption{Network architecture. 
For each convolution, the kernel size and the number of output channels are
indicated at the top and at the bottom, respectively
}
\label{fig:architecture}
\vspace{-1em}
\end{figure*}
More recently, with the spread of data-driven approaches,
a multitude of
local feature extraction
methods based on deep learning have emerged.
Early methods relied on existing keypoint detectors and focused on designing networks meant to extract the corresponding
descriptors \cite{hardnet, l2-net}.
Later, motivated by the tight entanglement of keypoints and descriptors, the focus shifted towards the design of
network architectures meant to predict keypoints and descriptors jointly \cite{d2-net, r2d2, aslfeat}.
Our method belongs to the latter group.

After the feature extraction step, 
in the pairwise matching,
the descriptors at the different images are compared
against each other in order to establish the 
correspondences.
This last step
is often responsible for a considerable part of the 
total computational cost of the sparse correspondence search.
In order to reduce the matching complexity, we propose 
a deep feature extraction network
capable of extracting multiple complementary keypoint sets.
This permits to restrict the comparison between the descriptors
at the different images only to the descriptors that belong to the same set,
thus reducing the matching computational complexity.
In order to train such a network, we propose a novel unsupervised loss
that discourages overlaps between the different keypoint sets.

For 3D reconstruction tasks, it has been shown that the keypoint distribution 
has a strong influence on the quality of the recovered camera poses \cite{down-to-earth}, 
which leads to worse results when large image portions are not covered.
To encourage an even keypoint distribution, 
we employ the unsupervised loss formulation originally proposed by \cite{r2d2}.
However, the use of this loss may lead to detections in
non-discriminative regions as well.
In order to mitigate this side effect, 
we propose a variance-based weighting scheme that dampens the loss in areas where the descriptors are less discriminative.

Differently from the classical methods, which rely on carefully handcrafted algorithms,
deep methods require large amounts of data in order to generalize 
to unseen scenes.
Moreover, depth maps and poses generated from 3D reconstructions, that are possibly inaccurate and incomplete, are often used for training \cite{r2d2, aslfeat}.
In an attempt to overcome these limitations, we train our feature extraction network
exclusively on images warped using random homographies. 
Furthermore, we augment the data with photometric distortions.

Our contribution is threefold:
\begin{itemize}
    \item We propose a deep architecture, named MD-Net, trained with a novel unsupervised loss formulation,
    which is capable of extracting multiple complementary sets of features. 
    This reduces the computational complexity of the subsequent matching phase. 
    \item A training loss re-weighting 
    based on the local
    descriptor variance is introduced. 
    This discourages the detection of 
    keypoints with less discriminative descriptors.
    \item Our feature extraction network, which is trained exclusively 
    on images warped using random homographies,
    generalizes well to 3D-related tasks as proven on two well known
    online benchmarks \cite{aachen, imb}.
\end{itemize}

\section{Related works}
In the last decades, a multitude of algorithms addressing the sparse correspondence problem have been designed:
in-depth evaluations have been carried out in 
\cite{localfeaturesurvey, localfeaturebenchmark0, localfeaturebenchmark1}.
With the advent of deep learning, data-driven methods were proposed to address one or more steps
of the existing feature extraction pipelines.
Early methods were trained to either detect repeatable keypoints \cite{tilde, taskdetector, quadnet, keynet},
or to distill compact descriptors from normalized patches,
previously extracted by means of a classical method \cite{hardnet, l2-net, sosnet}.
Later, deep methods were proposed to both detect keypoints and extract their descriptors \cite{lift, superpoint, lfnet},
with a shift towards joint learning with D2-Net, R2D2 and ASLFeat \cite{r2d2, d2-net, aslfeat}.
Differently from the already listed approaches,
\cite{disk} uses reinforcement learning to train a deep network 
for sparse feature extraction, 
obtaining good performance at the cost of a more expensive training procedure.
Our method is most closely related to R2D2 \cite{r2d2},
with which we share the core architecture and one of the unsupervised losses.
Differently from R2D2, 
we employ a variance-based loss dampening, 
supported by a two-stage training scheme,
to avoid detections in areas where
the resulting descriptors are not locally discriminative.
Additionally, our network is capable of detecting multiple complementary sets of keypoints.
While in \cite{largescale} a weight is predicted for each local features based on the relevance for the downstream image retrieval task,
our loss re-weighting is based on a parameter-free local measure of discriminativeness.

Deep learning has been applied successfully to the matching task as well, 
with \cite{superglue} completely replacing the traditional matching based on mutual nearest neighbors
and other methods proposing learnt outlier filters \cite{oanet, acne}.
These methods lead to better matching results, but increase the matching computational complexity significantly.

Multiple strategies have been proposed in order to reduce the
matching computational complexity for SfM pipelines.
These are particularly useful when dealing with the reconstruction of a scene
from an unordered set of images, 
potentially captured in different conditions.
In fact, in this scenario,
correspondences need to be established by matching all the possible image pairs,
which results in a complexity growing
quadratically in both the number of images and the number of extracted keypoints per image.
One possible approach toward reducing this computational burden 
is to lower the number of image pairs
by using strategies based on image similarities \cite{schoenberger2016vote}.
Alternatively, the number of matching operations can be reduced 
by using approximate nearest neighbor algorithms \cite{approximate_nearest_neighbor, flann}.
However, the former approach introduces the risk of missing valid image pairs
and the latter decreases the quality of the matches \cite{imb}.
For those reasons, when high reconstruction quality is required,
many 3D reconstruction pipelines still match all the possible image pairs
and use the exact Mutual Nearest Neighbor (MNN) matching
\cite{schoenberger2016sfm, opensfm}.
With MD-Net, we propose a novel approach that reduces the matching complexity
by extracting a predefined number of disjointed feature sets at each image, 
which permits to limit the matching to the sole features in the same set.

\section{Model overview}

\subsection{Network architecture}
The network architecture, depicted in Fig.~\ref{fig:architecture}, is a streamlined version of R2D2 \cite{r2d2}
with the addition of our multi-detector branch.
The backbone consists of a fully convolutional network where the commonly used convolution pyramid is replaced
by a series of dilated convolutions, meant to increase the effective field-of-view of the network without
lowering the output resolution.
The backbone processes the input RGB image 
\mbox{$\bm{I} \! \in \! \mathbb{R}^{3 \times H \times W} \!$ }
and outputs the feature volume 
\mbox{$\bm{F} \! \in \! \mathbb{R}^{C \times H \times W} \!$}.
The feature volume is then fed to two different branches: the descriptor branch and the multi-detector branch.
In the descriptor branch the feature volume is L2-normalized along the channel dimension 
to produce the final descriptor volume 
\mbox{$\bm{V} \! \in \! \mathbb{R}^{C \times H \times W} \!$}.
This associates a $C$-dimensional descriptor vector to each pixel of the input image.
In the multi-detector branch, instead, the feature volume $\bm{F}$ is squared and a single 1x1 convolutional layer
is used to generate a detection heatmap volume 
\mbox{$\bm{D} \! \in \! \mathbb{R}^{N \times H \times W} \!$}
where $N$ is the desired number of keypoint sets.
In fact, each channel of this volume, hereafter referred as 
\mbox{$\bm{D}^n \! \in \! \mathbb{R}^{H \times W} \!$}
with  
\mbox{$n\! =\! 0, 1, \ldots, N\!-\!1$}, 
will be used to extract one set of keypoints.
The resulting Multi-Detector network, named \mbox{MD-Net}, is rather compact and counts less than half a  million parameters.

\begin{figure}
\centering
\vspace*{0.5ex}
\includegraphics[width=\linewidth]{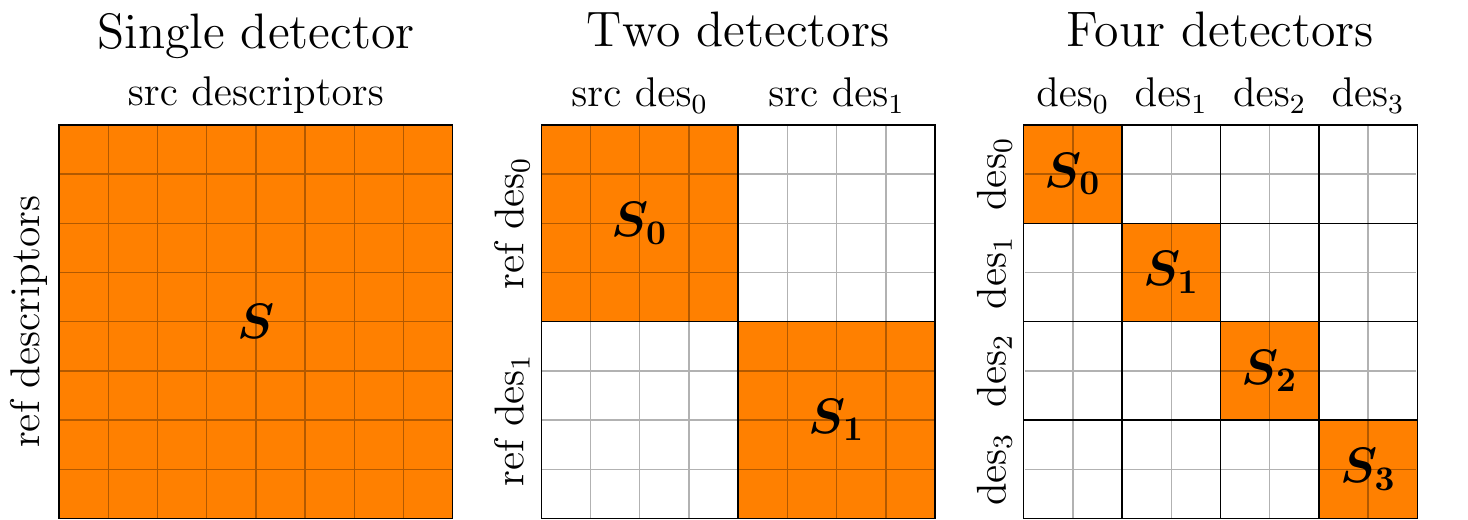}
\caption{
Graphical representation of the computational complexity 
associated with the
mutual nearest neighbors descriptor matching.
While in the \textit{single-detector} case
the $S$ matrix is computed for each possible descriptor combination,
in the \textit{multi-detector} cases 
it is sufficient to compute the smaller $S_n$
between the different sets of descriptors $\text{des}_n$ with $n \in \{0, \ldots, N-1 \}$.
The computational cost, shown in orange,
results halved in the case of \textit{two detectors} 
and reduced by a factor of 4 in the case of \textit{four detectors}.
}
\label{fig:multidet_complexity}
\vspace{-1em}

\end{figure}

\subsection{Feature Extraction and Matching}
\label{subsec:feature_extraction_and_matching}

At each heatmap $\bm{D}^n$, the candidate keypoints are detected as the pixel coordinates of the heatmap local maxima,
after filtering out low values and applying a local Non Maxima Suppression (NMS).
Given a budget of $M$ keypoints, for each heatmap we select only the $M/N$ candidate keypoints with the highest values in the
heatmap.
Finally, we obtain the local features by coupling each keypoint with its descriptor, 
sampled from the descriptor volume $\bm{V}$ at the keypoint pixel location.

For a pair of images with $M$ features each, the Mutual Nearest Neighbor matching boils down to computing a
distance matrix $\bm{S} \in \mathbb{R}^{M \times M}$
between the two image descriptor sets, which has a computational complexity
$\mathcal{O}(M^2)$.
Thanks to our network architecture instead, only descriptors associated 
with the same detector heatmap need to be matched,
which reduces the distance matrix size to $M/N \times M/N$ and the corresponding computational complexity
for each set to $\mathcal{O}(M^2/N^2)$.
Repeating the matching for each one of the $N$ feature sets results in an overall complexity reduction factor $N$, as follows:
\begin{equation}
    \mathcal{O}_\text{N-sets} = 
    N \cdot \mathcal{O}\bigg(\frac{M^2}{N^2}\bigg) = 
    \mathcal{O}\bigg(\frac{M^2}{N}\bigg)
\end{equation}
A visual intuition for the reduced computational complexity is provided in Fig.~\ref{fig:multidet_complexity}.
The aggregated matches are obtained joining all the sets of matches.

\section{Loss formulation}
The loss formulation can be split in two main components: 
the \textit{descriptor loss} and \textit{detector loss}, applied at the output of the corresponding branches, respectively.

\subsection{Descriptor loss}

The \textit{descriptor loss} goal is to promote discriminative descriptors, that permit to
recognize the correct correspondences between the keypoints of two images.
Similarly to previous works \cite{l2-net, hardnet}, we frame descriptor learning as a metric learning problem,
where we promote that two corresponding keypoints have similar descriptors, while non corresponding keypoints
should have dissimilar ones.
To this purpose, we employ a simple hinged formulation of the Triplet Loss:
\begin{equation}
   \mathcal{L}_{Triplet} = \mean_{t \in \mathcal{T}} \big( \max(0, m - \bm{v_a}^t \cdot \bm{v_p}^t + \bm{v_a}^t \cdot \bm{v_n}^t) \big)
   \label{eq:triplet}
\end{equation}
where $\cdot$ denotes the inner product, $\mathcal{T}$ is the set of all the sampled triplets,
$m$ is the hinge margin and $\bm{v_a}^t$, $\bm{v_p}^t$, $\bm{v_n}^t \! \in \! \mathbb{R}^C$ refer to 
the \textit{anchor} descriptor, the \textit{positive} correspondence descriptor and
one \textit{negative} descriptor, respectively.
While it is trivial to build the \textit{(anchor, positive)} descriptor pair,
if the geometric transformation 
\mbox{$g: \mathbb{R}^2 \mapsto \mathbb{R}^2$} that relates the considered image pair is known,
there is a virtually infinite number of possible \textit{(anchor, negative)} candidates. 
As suggested in \cite{hardnet}, we pick the \textit{negative} following the \textit{Hardest-in-Batch} strategy.

\subsection{Detector loss}
The \textit{detector loss} goal is twofold.
First, promoting heatmaps with well localized maxima, as these will determine the detected keypoints.
Second, promoting repeatable heatmaps: content appearing in two images should lead to similar heatmaps,
such that keypoint correspondences can be established between the two images.
We design our loss as the sum of three components:
the \textit{peakyness loss}, the \textit{similarity loss} and the \textit{dissimilarity loss}.
While the first two losses are applied to all the detection heatmaps in $\bm{D}$ independently
and then mean aggregated,
the \textit{dissimilarity} loss formulation considers each possible pair of detection heatmaps,
in order to discourage any overlap between sets of keypoints selected by different detectors.
For the sake of clarity, 
in the following we express each loss for the single pixel $(i, j)$.
The losses are then mean aggregated over the entire 
$H \times W$ image domain.

\subsubsection{Peakyness loss}
\label{sec:peaky}
In order to encourage the network to produce well distributed local peaks,
while avoiding non-discriminative areas,
we propose a modified version of the \textit{peaky} loss formulated in \cite{r2d2}. 
The loss is defined as follows:
\begin{equation}
    \mathcal{L}_\text{peaky}(\bm{D}^n)_{ij} = 
    W_{ij} 
    \left(1 -
    \left( 
        \max_{kl \in \mathcal{P}_{ij}} \bm{D}^n_{kl} -
        \mean_{kl \in \mathcal{P}_{ij}} \bm{D}^n_{kl}
    \right)
    \right)
    \label{eq:peaky_loss}
\end{equation}
where $\mathcal{P}_{ij}$ is a square patch centered at the pixel $(i, j)$
and $W_{ij}$ is a weight designed to avoid peaks in 
areas where local descriptors are not discriminative
defined as follows:
\begin{equation}
    W_{ij} = 
    \mean_{c \in \mathcal{C}} 
    \raisebox{-1pt}{\bigg(\bigg(}     
    \mean_{pq \in \mathcal{B}_{ij}} \bm{F}_{pq}^2 - 
    \bigg(\mean_{pq \in \mathcal{B}_{ij}} \bm{F}_{pq} \bigg)^{\!\!2}
    \,\bigg)_{\!\!c}\,\bigg)
    \label{eq:variance}
\end{equation}
and it represents the local variance of the backbone output $\bm{F}$,
computed over a patch $\mathcal{B}_{ij}$ centered at $(i, j)$, averaged along the channel dimension.
Additionally, the loss in Eq.\eqref{eq:peaky_loss} is computed on the detection heatmaps in
$\roverline{\bm{D}} \in \mathbb{R}^{N \times H \times W}$
obtained from the warped image $g(\bm{I}) \in \mathbb{R}^{H \times W}$.
The two losses are averaged.
An example of the effect of the pixelwise weighting is shown in Fig.~\ref{fig:peakyV},
where the detection heatmap appears smoother in the less discriminative regions.

\subsubsection{Similarity loss}
In order to promote repeatable heatmaps, we adopt the following loss, that
enforces consistency between the heatmaps produced by $\bm{I}$ and $g(\bm{I})$:
\begin{equation}
    \mathcal{L}_\text{sim}(\bm{D}^n, \roverline{\bm{D}}^{n})_{ij} = 
    \Big(\bm{D}_{ij}^n - g^{-1}\Big(\roverline{\bm{D}}^{n}\Big)_{ij}\Big)^2
    \label{eq:similarity_loss}
\end{equation}
where $g^{-1}(\cdot)$ denotes the inverse warping.

\begingroup
\setlength{\tabcolsep}{1 pt}
\renewcommand{\arraystretch}{0.6}
\begin{figure}
\vspace*{0.8ex}
\centering
\begin{tabular}{cccc}
\includegraphics[width=0.24\linewidth, trim={1.0cm 4.0cm 7.0cm 0.11cm}, clip]{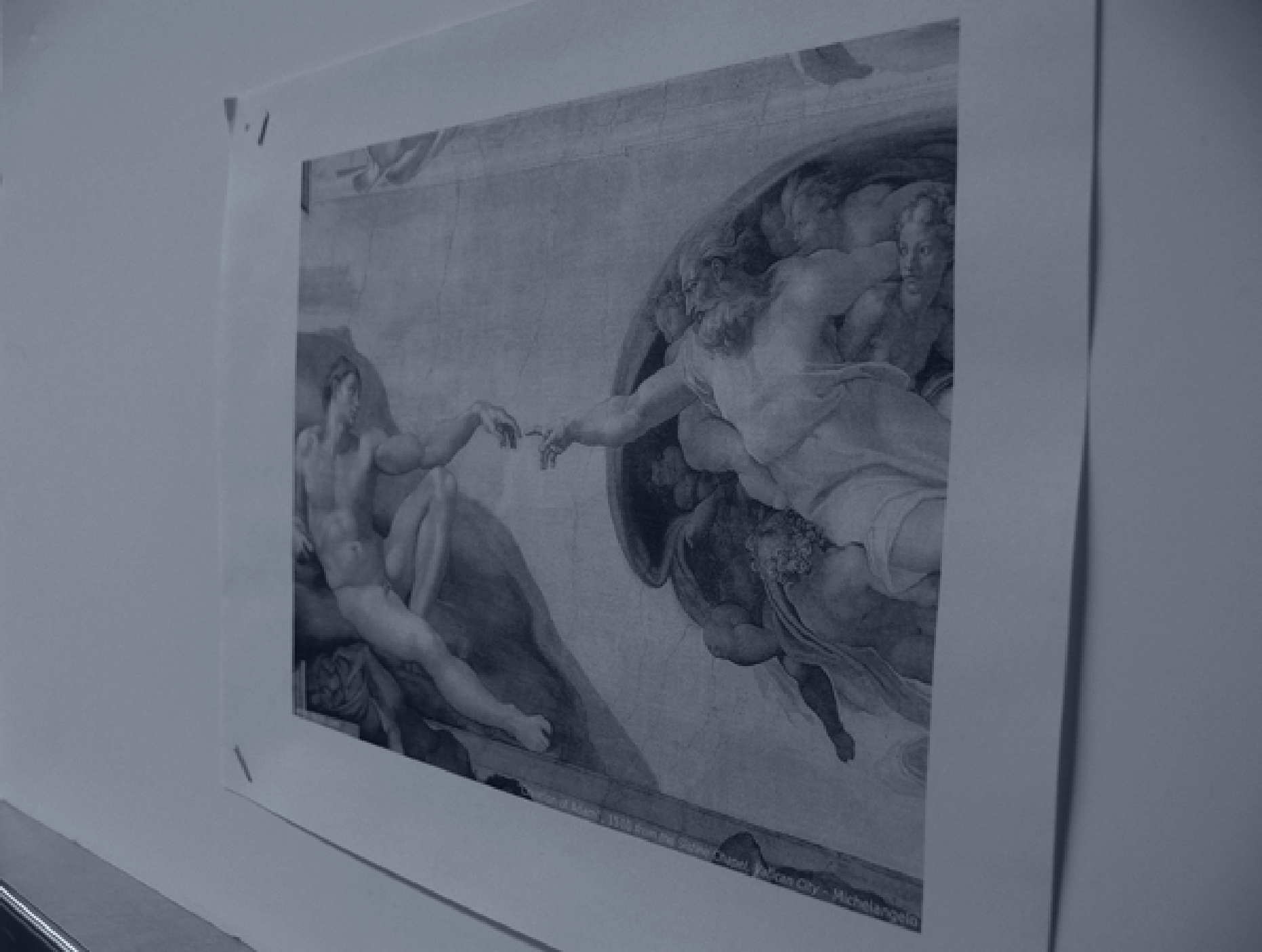} &
\includegraphics[width=0.24\linewidth, trim={1.0cm 4.0cm 7.0cm 0.0cm}, clip]{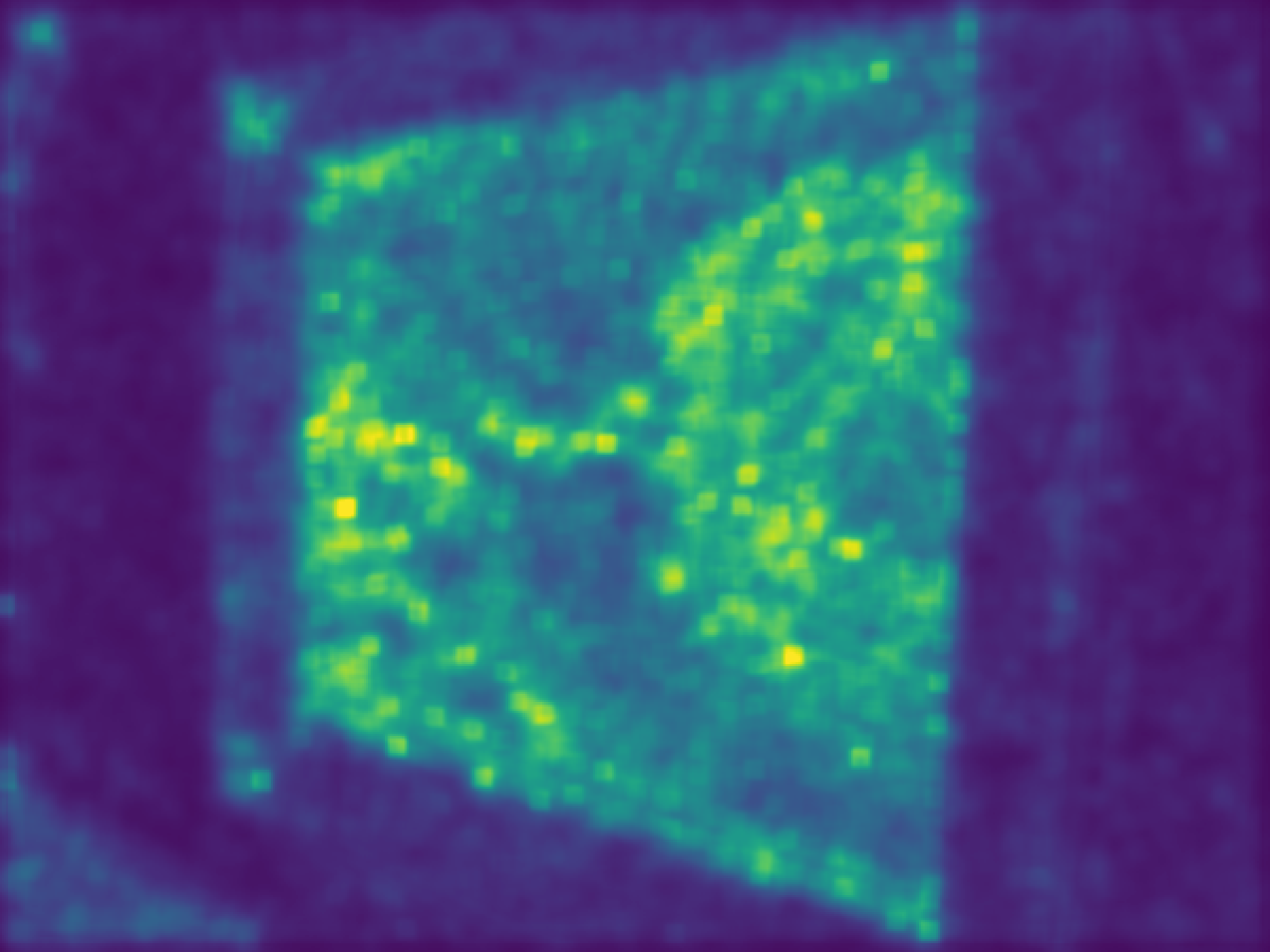} &
\includegraphics[width=0.24\linewidth, trim={1.0cm 4.0cm 7.0cm 0.0cm}, clip]{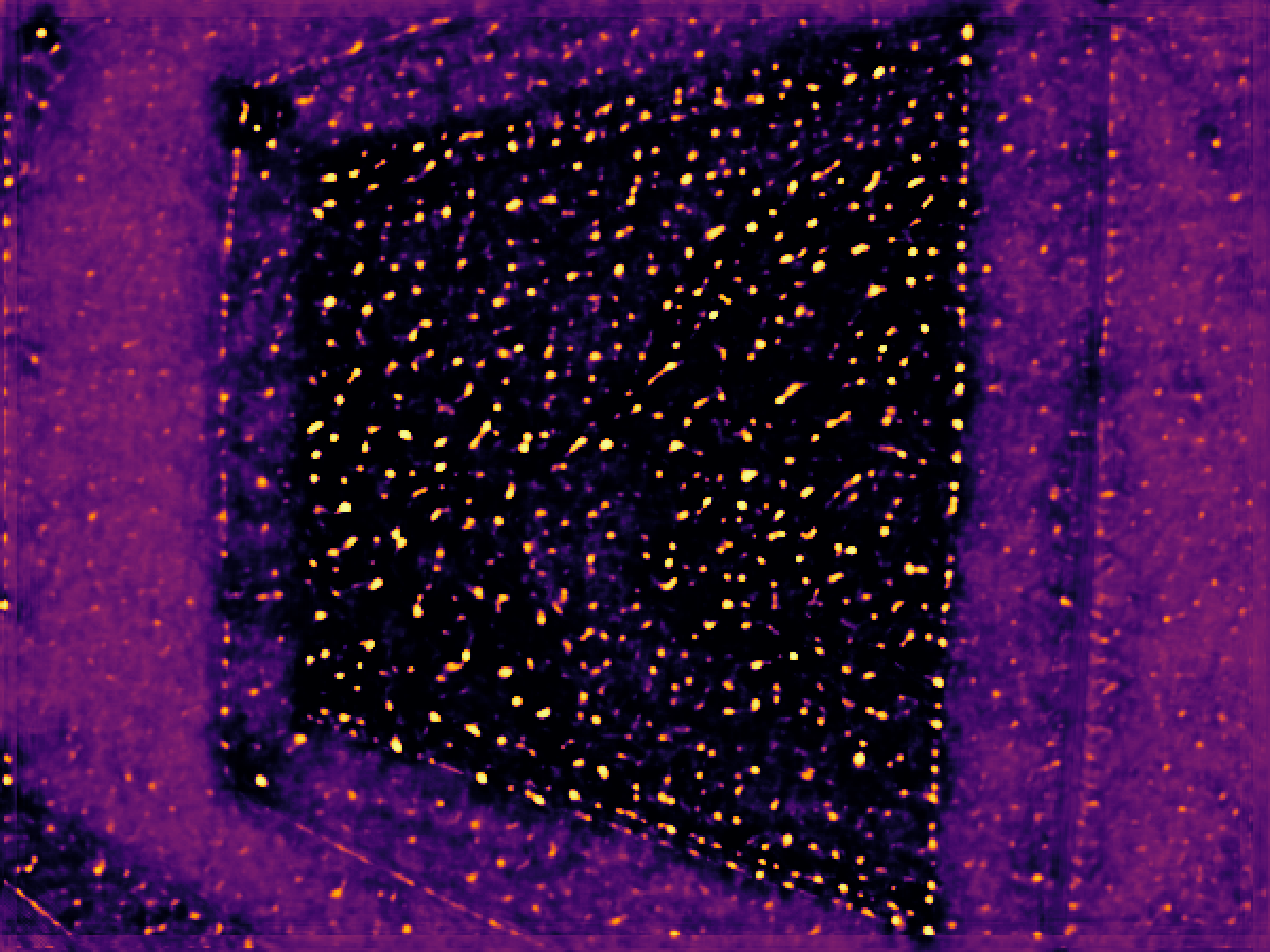} &
\includegraphics[width=0.24\linewidth, trim={1.0cm 4.0cm 7.0cm 0.0cm}, clip]{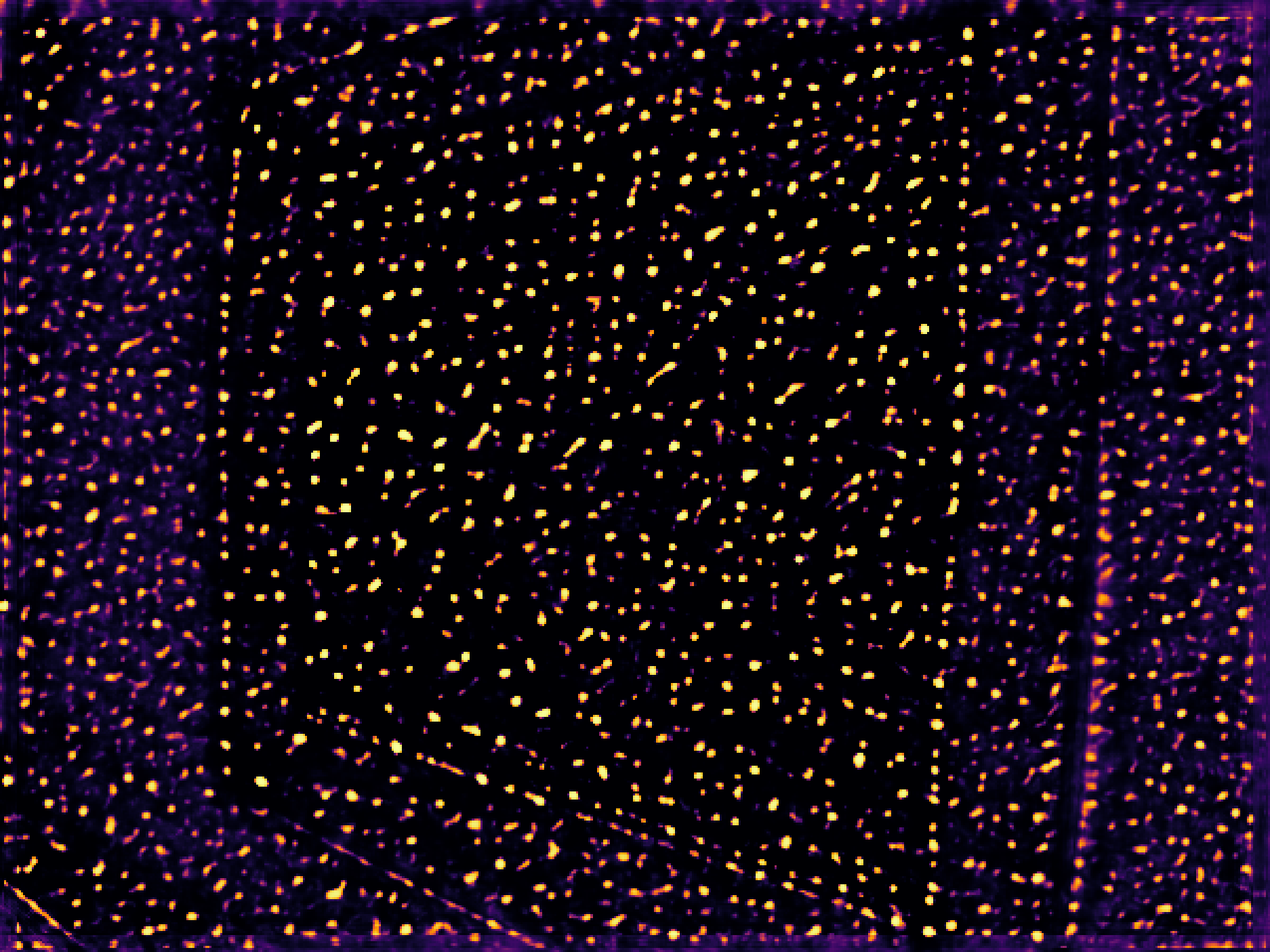}
\end{tabular}
\caption{
Qualitative evaluation of the \textit{peakyness} loss variance-based weighting.
On the left, the source image and the computed local descriptors variance. 
On the right, the detection heatmaps obtained when training with and without our
variance weighting in Eq.~\ref{eq:peaky_loss}, respectively.
Our variance-based weighting scheme smooths out the peaks in the
non-discriminative region at the left border of the image,
thus avoiding detections there.
}
\label{fig:peakyV}
\vspace{-1em}
\end{figure}
\endgroup

\subsubsection{Dissimilarity loss}

Finally, in order to promote that the $N$ detection heatmaps in $\bm{D}$ lead to different sets
of keypoints, we propose a novel loss that penalizes co-located peaks for each pair $(\bm{D}^m, \bm{D}^n)$.
Our loss is formulated as follows:
\begin{equation}
    \mathcal{L}_\text{dissim}(\bm{D}^0, \cdots, \bm{D}^N)_{ij} = \binom{N}{2}^{-1} \sum_{\mathclap{\substack{~~~0 \le n < N-1 \\ n < m < N}}} ~ \bm{D}^m_{ij}  \bm{D}^n_{ij}
    \label{eq:dissimilarity_loss}
\end{equation}
where N is the number of detectors and the binomial $\binom{N}{2}$ is the number of possible detector heatmap combinations.
Similarly to the \textit{peakyness} loss, Eq.~\eqref{eq:dissimilarity_loss} is applied to the $N$ detection heatmaps
in $\roverline{\bm{D}}$ as well.
Fig.~\ref{fig:synth-multidet} provides an example of the keypoint sets obtained for $N=2$.

\section{Model training}

\begingroup
\setlength{\tabcolsep}{1 pt}
\renewcommand{\arraystretch}{0.6}
\begin{figure}
\vspace*{0.8ex}
\centering
\begin{tabular}{cccc}
\includegraphics[width=0.23\linewidth, trim={0.7cm 0.7cm 0.7cm 0.7cm}, clip]{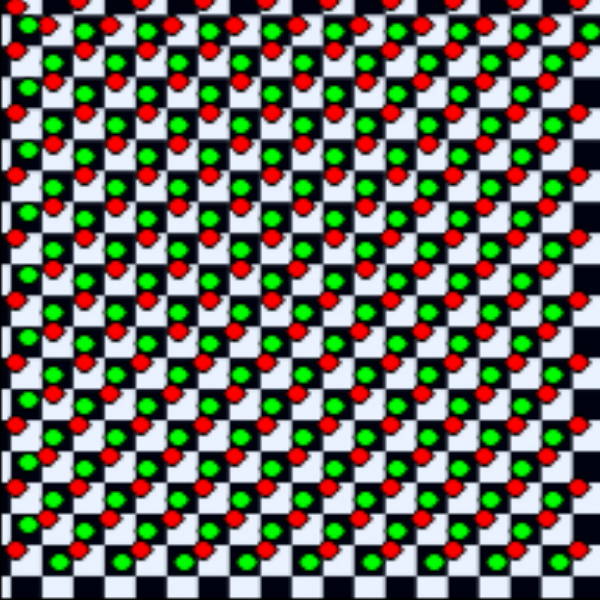}
&  \includegraphics[width=0.23\linewidth, trim={0.7cm 0.7cm 0.7cm 0.7cm}, clip]{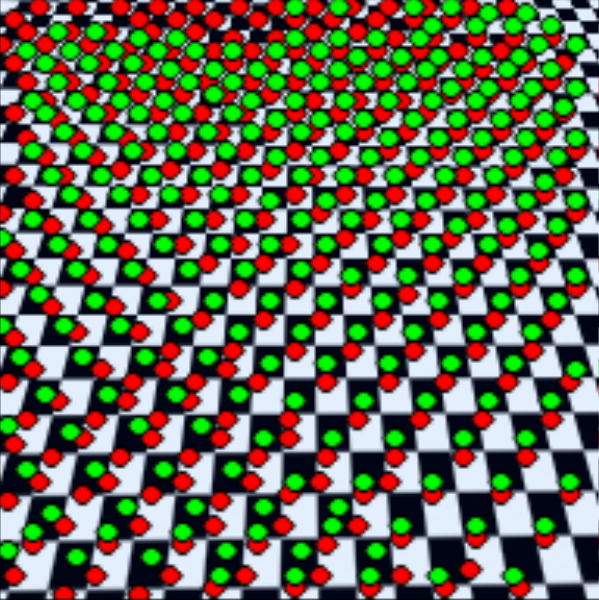} 
& \hspace{0.2cm}  \includegraphics[width=0.23\linewidth, trim={1.2cm 1.2cm 1.2cm 1.2cm}, clip]{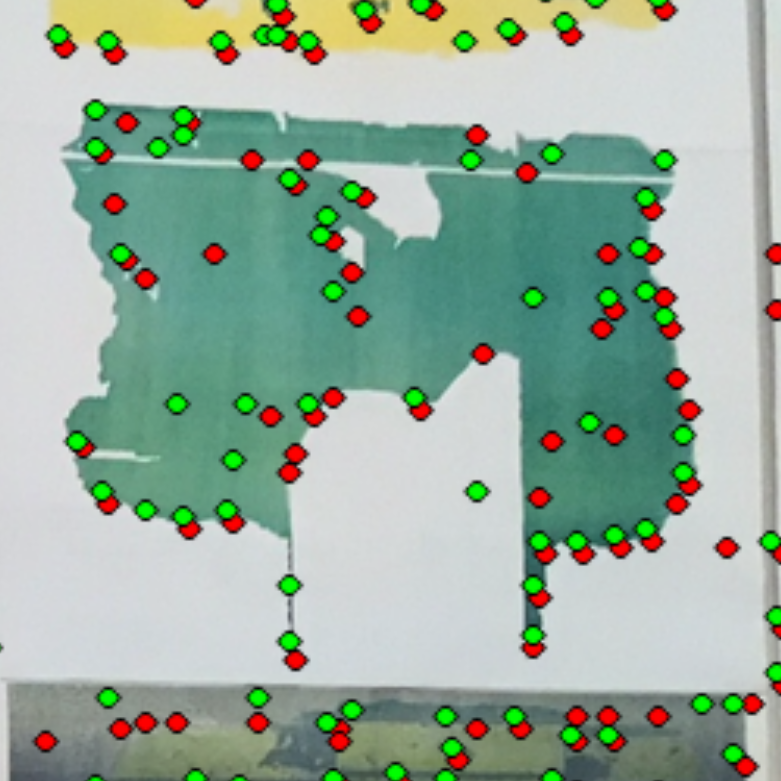}
&  \includegraphics[width=0.23\linewidth, trim={1.2cm 1.2cm 1.2cm 1.2cm}, clip]{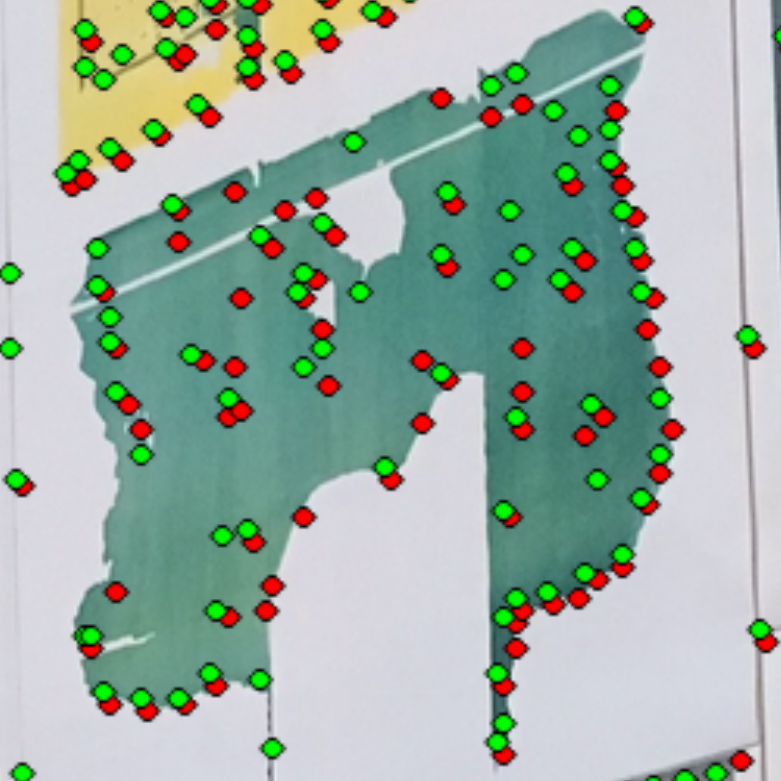}
\end{tabular}
\caption{
Example images showing the complementarity of the keypoints sets detected by our architecture with
2 detectors.
The two sets of keypoints are represented in red and green.
On the left, a pair of checkerboard images: 
the red set keypoints are extracted
always at the white squares, the green set ones at the black squares.
It's worth noticing that the network has never seen checkerboard patterns during the training.
On the right, another example pair.
}

\label{fig:synth-multidet}
\vspace{-1em}
\end{figure}
\endgroup

While the local variance computed over the input image can
be helpful in discerning 
textured and flat areas, it does not directly relate to the local
descriptor discriminativeness.
For this reason, Eq.~\eqref{eq:peaky_loss} employs the backbone output local variance instead, which is
related to the descriptor volume directly and it is therefore more suitable to avoid keypoint detections in
areas whose descriptors would not be particularly discriminative.
However, this reasoning does not hold true at the beginning of training,
when the network weights are randomly distributed and the predicted descriptor volume is not meaningful.
Thus, we adopt a \textit{two-stage training} procedure:
\begin{enumerate}
    \item First, in the \textit{descriptor volume priming}, we train only the backbone and the descriptor branch with the loss $\mathcal{L} = \mathcal{L}_\text{triplet}$.
    \item Then, in the \textit{joint training}, we train our overall architecture with $\mathcal{L} = \mathcal{L}_{triplet} + \alpha \mathcal{L}_{peaky} + \beta \mathcal{L}_{sim} + \gamma \mathcal{L}_{dissim}$ where $\alpha$, $\beta$ and $\gamma$ balance the individual losses.
\end{enumerate}

The descriptor volume priming represents the main training effort, while the joint training needs only few iterations.
An added benefit of this training procedure is that
changing the number of keypoint sets $N$ requires us to repeat only the joint training stage.
Finally, during the joint training, the local variance $W$ is used purely as a weighting term,
i.e., the weight gradients do not participate in the backpropagation.

\section{Experiments}
\label{section:experiments}

\subsection{Training details}

We train MD-Net on $192 \times 192$ patches randomly drawn from
the \textit{Revisiting Oxford and Paris distractors} dataset \cite{oxford_paris}.
We implement our model in PyTorch \cite{pytorch} and train it with the Adam optimizer \cite{adam} ($\beta_1=0.9$, $\beta_2=0.999$)
and fixed $\text{lr}\!=\!1e^{-4}$ on a single Nvidia GTX1080Ti.
The descriptor volume priming consists of $70k$ iterations and takes $13$ hours.
Instead, the joint training consists only of $1k$ iterations and is completed in $12$ minutes.
Each iteration employs a batch of $10$ patches.
Overall, the training procedure consumes a total of 710k images.
Concerning the descriptor loss in Eq.~\eqref{eq:triplet},
we set the hinge margin $m=1$,
sample the \textit{positive} and \textit{negative} descriptors on a regular grid with step 10px
and classify a descriptor as negative candidate when it is more than $5\text{px}$ away from the correct location.
We adopt 128-dimensional descriptors.
Concerning the peaky loss in Eqs.~\eqref{eq:peaky_loss} and \eqref{eq:variance}, we set $\mathcal{P}_{ij}$ and $\mathcal{B}_{ij}$
to be $17 \times 17$ and $9 \times 9$ patches, respectively.
Finally, for the training scenario with $N=2$ detectors, we use the loss weights $\alpha=1.0$, $\beta=4.0$ and $\gamma=0.5$.

\begingroup
\begin{table}[!t]
\vspace*{0.5ex}
\setlength{\tabcolsep}{4 pt}
\renewcommand{\arraystretch}{1.2}
\renewcommand\cellalign{cc}
\caption{Ablation studies on HPatches - overall}
\vspace*{-1.2ex}
\label{tab:hpatches_ablations}
\centering
\begin{tabular}{|l|c|c|c|c|c|c|}
\hline
\multicolumn{1}{|c|}{\multirow{2}{*}{Method}} & \multicolumn{3}{c|}{MMA $\uparrow$} & \multicolumn{3}{c|}{MS $\uparrow$} \\
\cline{2-7}
& @1px & @2p & @3px & @1px & @2px & @3px \\
\hline
MD-1-Net No-Var & 0.385  &     0.622  & 0.740      &     0.161  &      0.255  &     0.298   \\
\hline
MD-1-Net &    \bf{0.398} & \bf{0.638} & \bf{0.757} &     0.196  &      0.306  &     0.356   \\
MD-2-Net &    \bf{0.398} & \uf{0.630} & \uf{0.743} & \uf{0.206} & \uf{0.317}  & \bf{0.369}  \\
MD-4-Net &    \bf{0.398} &     0.621  &     0.731  & \bf{0.210} & \bf{0.319}  & \bf{0.369}  \\
MD-8-Net &        0.346  &     0.541  &     0.636  &     0.184  &     0.280   &     0.325   \\
\hline
\end{tabular}
\end{table}
\endgroup

\begingroup
\begin{table}[!t]
\vspace*{0.5ex}
\setlength{\tabcolsep}{4 pt}
\renewcommand{\arraystretch}{1.2}
\renewcommand\cellalign{cc}
\caption{Comparison on HPatches}
\vspace*{-1.2ex}
\label{tab:hpatches}
\centering
\begin{tabular}{|l|c|c|c|c|c|c|c|}
\cline{2-8}
\multicolumn{1}{c|}{} & \multicolumn{1}{c|}{\multirow{2}{*}{Method}} & \multicolumn{3}{c|}{MMA $\uparrow$} & \multicolumn{3}{c|}{MS $\uparrow$} \\
\cline{3-8}
\multicolumn{1}{c|}{} & \multicolumn{1}{c|}{} & @1px & @2p & @3px & @1px & @2px & @3px \\
\cline{2-8}
\noalign{\vskip\doublerulesep}
\hline
\multirow{4}{*}{\rotatebox[origin=c]{90}{v}} 
& MD-2-Net (ours)          & \uf{0.316} & \bf{0.600} & \bf{0.722} & \uf{0.171} & \uf{0.313} & \uf{0.393} \\
& R2D2 \cite{r2d2}         &     0.280  & \uf{0.568} & \uf{0.700} &     0.118  &     0.228  &     0.273  \\
& ASLFeat \cite{aslfeat}   & \bf{0.332} &     0.565  &     0.675  & \bf{0.203} & \bf{0.338} & \bf{0.398} \\
& Upright-SIFT \cite{sift} &     0.313  &     0.472  &     0.533  &     0.167  &     0.247  &     0.277  \\
\hline\hline
\multirow{4}{*}{\rotatebox[origin=c]{90}{i}} 
& MD-2-Net (ours)          & \bf{0.480} &     0.658  &     0.765  & \uf{0.242} & \uf{0.323} & \uf{0.368} \\
& R2D2 \cite{r2d2}         &     0.377  & \uf{0.660} & \bf{0.797} &     0.170  &     0.285  &     0.336  \\
& ASLFeat \cite{aslfeat}   & \uf{0.469} & \bf{0.664} & \uf{0.774} & \bf{0.290} & \bf{0.398} & \bf{0.456} \\
& Upright-SIFT \cite{sift} &     0.344  &     0.475  &     0.528  &     0.161  &     0.216  &     0.238  \\
\hline\hline
\multirow{4}{*}{\rotatebox[origin=c]{90}{overall}} 
& MD-2-Net (ours)          & \bf{0.398} & \bf{0.630} & \uf{0.743} & \uf{0.206} & \uf{0.317} & \uf{0.369} \\
& R2D2 \cite{r2d2}         &     0.326  &     0.612  & \bf{0.747} &     0.143  &     0.255  &      0.304 \\
& ASLFeat \cite{aslfeat}   & \bf{0.398} & \uf{0.613} &     0.723  & \bf{0.245} & \bf{0.367} & \bf{0.426} \\
& Upright-SIFT \cite{sift} &     0.327  &     0.473  &     0.531  &     0.164  &     0.232  &      0.258 \\
\hline
\end{tabular}
\vspace{-1.5em}
\end{table}
\endgroup

\subsection{Benchmarks}

We test MD-Net on three popular benchmarks: \textit{HPatches} \cite{hpatches}, \textit{Aachen} day-night \cite{aachen_extended}
and the \textit{Image Matching Benchmark} \cite{imb}.
In all the experiments we employ MD-Net with $N = 2$ detectors, denoted MD-2-Net.
The filtering threshold and the NMS radius introduced for the keypoint extraction in Sec.~\ref{subsec:feature_extraction_and_matching} are set to 0.7 and $3\text{px}$,
respectively.
We run \mbox{MD-2-Net} on a multi-scale image pyramid obtained by down scaling the input image by a factor $\sqrt 2$ until the shortest image dimension drops below $256\text{px}$.
Finally, for each detector, we select the $M/N$ keypoints with the highest scores across the multiple scales.
The main metrics in the experiments are the following, involving a pair of images
that have to be matched:
\begin{itemize}
    \item \textbf{MMA}: The \textit{Mean Matching Accuracy}
    is the mean ratio between the number of correct matches and the total number of proposed matches \cite{d2-net}.
    \item \textbf{MS}: The \textit{Matching Score}
    is the mean ratio between the number of correct matches and the number of keypoints extracted at one image
    in the area shared with the other.
    The metric is computed for both the images and the results are averaged \cite{superpoint}.
    \item \textbf{mAA}: The \textit{mean Average Accuracy}
    is the area under the curve of the fraction of correctly estimated relative poses as a function of the 
    pose error \cite{imb}.
\end{itemize}
MMA and MS are evaluated at a given pixel error threshold.
In all the tables we 
represent the best result in bold and we underline the second best.
We compare MD-2-Net with two state-of-the-art deep feature extraction networks: R2D2 \cite{r2d2} and ASLFeat \cite{aslfeat}.
We employ their official implementations and adopt either their default parameters or those specified by the authors
for each benchmark, when provided.
In addition, 
we consider also Upright-SIFT \cite{sift}, 
the baseline method in the Image Matching Benchmark \cite{imb}, 
employing their implementation.
For the purpose of a fair comparison, we do not compare with methods employing deep matchers, such as \cite{superglue}.

\begingroup
\begin{table}[!t]
\caption{Aachen day-night visual localization v1.1}
\vspace*{-1.2ex}
\label{tab:aachen}
\renewcommand{\arraystretch}{1.2}
\centering
\begin{tabular}{|c|c|c|c|c|}
\cline{2-5}
\multicolumn{1}{c|}{} & \multirow{2}{*}{Method} & \multicolumn{3}{c|}{successfully localized percentage $\uparrow$} \\
\cline{3-5}
\multicolumn{1}{c|}{} & & 0.25m, 2° & 0.5m, 5° & 5m, 10°\\
\cline{2-5}
\noalign{\vskip\doublerulesep}
\hline
\multirow{4}{*}{\rotatebox[origin=c]{90}{8k kpts}} 
& MD-2-net (ours)          & \bf{70.2} & \uf{83.2} & \uf{96.3} \\
& R2D2 \cite{r2d2}         &     66.0  &     82.2  &     94.8\\
& ASLFeat \cite{aslfeat}   & \uf{69.6} & \bf{84.8} & \bf{97.4}\\
& Upright-SIFT \cite{sift} &     54.5  &     69.6  &     79.1\\
\hline\hline
\multirow{4}{*}{\rotatebox[origin=c]{90}{20k kpts}} 
& MD-2-net (ours)          & \uf{69.1} & \uf{84.8} & \bf{97.9}\\
& R2D2 \cite{r2d2}         &     68.1  &     83.8      & 96.9\\
& ASLFeat \cite{aslfeat}   & \bf{72.3} & \bf{86.4} & \bf{97.9}\\
& Upright-SIFT \cite{sift} &     62.3  &     78.5  &     90.1 \\
\hline
\end{tabular}
\vspace{-1.5em}
\end{table}
\endgroup

\subsubsection{HPatches \cite{hpatches}}
This benchmark considers both indoor and outdoor scenes
divided in two sets:
\textit{v} contains images of mostly planar scenes captured from different angles in the same lighting conditions,
while \textit{i} contains images captured from a fixed camera in different lighting conditions.
Each scene, 
$56$ for the $v$ set and $52$ for $i$, 
contains $6$ images
and the ground truth homographies linking the first image to all the others.
We evaluate following the methodology of D2-net \cite{d2-net} with a
maximum budget of $5$k keypoints per image and evaluation on the sets \textit{v}, \textit{i} and their union, denoted \textit{overall}.
The performance at error thresholds greater than a few pixels are of little interest
in real world applications, such as in 3D reconstruction, due to the tight geometric filters employed.
For this reason, in Tab.~\ref{tab:hpatches} we report only the numerical values of MMA and MS up to 3px error. 
\mbox{MD-2-Net} obtains competitive MMA results on all the three image sets, at all the error thresholds.
In particular, it is the best performing method in the \textit{overall} set at both 1px and 2px,
while following R2D2 closely at 3px.
Additionally, MD-2-Net provides good MS results, following the best performing method ASLFeat.

\begingroup
\begin{table*}[!t]
\vspace*{0.5ex}
\setlength{\tabcolsep}{4 pt}
\renewcommand{\arraystretch}{1.23}
\caption{Image matching benchmark - restricted keypoints 2048}
\vspace*{-1.2ex}
\label{tab:imb}
\centering
\begin{tabular}{|c|c|c|c|c|c|c|c|c|c|c|c|c|}
\cline{3-13}
\multicolumn{1}{c}{} & \multicolumn{1}{c|}{} & \multicolumn{5}{c|}{Stereo} & \multicolumn{5}{c|}{Multiview} & Avg \\
\cline{2-13}
\multicolumn{1}{c|}{} & Method & NF & NI $\uparrow$ & Rep@3px $\uparrow$ & MS@3px $\uparrow$ & mAA@10° $\uparrow$ & NM $\uparrow$ & NL $\uparrow$ & TL $\uparrow$ & ATE $\downarrow$ & mAA@10° $\uparrow$ & mAA@10° $\uparrow$\\
\cline{2-13}
\noalign{\vskip\doublerulesep}
\hline
\multirow{4}{*}{\rotatebox[origin=c]{90}{Phototourism}} 
& MD-2-net (ours) &     2047.5 & \bf{233.0} & 0.396      & \bf{0.792} & \bf{0.455} & \uf{238.6} & \bf{1391.5} & \bf{4.604} & \bf{0.411} & \bf{0.708} & \bf{0.581} \\
& R2D2 \cite{r2d2} &         2048.0 & \uf{201.5} & \uf{0.429} & 0.746      & \uf{0.390} & \bf{294.3} & \uf{1225.9} & 4.280      & \uf{0.478} & \uf{0.640} & \uf{0.515} \\
& ASLfeat \cite{aslfeat} &      2042.6 &      126.0 & \bf{0.431} & 0.749      & 0.337      & 157.5      & 1106.6      & \uf{4.415} & 0.533      & 0.556      & 0.446 \\
& Upright-SIFT \cite{sift} & 1892.8 &      98.6 & 0.333       & \uf{0.788} & 0.383      & 148.0      & 1165.7      & 4.118      & 0.524      & 0.555      & 0.469 \\
\hline\hline
\multirow{4}{*}{\rotatebox[origin=c]{90}{PragueParks}} 
& MD-2-net (ours) &     2048.0 & \bf{175.5} & 0.039      & \uf{0.027} & \bf{0.542} & \uf{236.3} &  \bf{605.8} & \bf{3.197} & 6.753      & \bf{0.451} & \bf{0.497} \\ 
& R2D2 \cite{r2d2} &         2048.0 & \uf{167.0} & 0.032      & 0.025      & \uf{0.539} & \bf{338.9} &  526.0      & \uf{3.170} & 6.837      & \uf{0.444} & \uf{0.491} \\
& ASLfeat \cite{aslfeat} &      2048.0 &      110.5 & \uf{0.059} & \bf{0.029} & 0.401      & 217.1      &  \uf{574.4} & 3.036      & \uf{6.414} & 0.400      & 0.403 \\
& Upright-SIFT \cite{sift} & 2048.0 &      119.8 & \bf{0.060} & \uf{0.027} & 0.414      & 157.3      &  433.3      & 2.989      & \bf{5.666} & 0.361      & 0.387 \\
\hline
\end{tabular}
\end{table*}
\endgroup

\subsubsection{Aachen Day-Night \cite{aachen, aachen_extended}}
This online benchmark is part of the long-term visual localization benchmark \cite{visuallocalization_benchmark}.
It consists of two sets of images of the German city Aachen.
The first set is captured during daytime and the corresponding ground truth camera intrinsics and poses are provided.
The second set is captured at night instead and the benchmark target is to re-localize these query images using the first set.
The online benchmark has been recently updated to v$1.1$ with more precise ground truth poses and additional query images. 
For a fair comparison, we run MD-2-Net, R2D2, ASLFeat and UprightSIFT using the same re-localization pipeline based on COLMAP \cite{schoenberger2016sfm}, available at \cite{visuallocalization_code}. 
The results are reported in Tab~\ref{tab:aachen}, where MD-2-Net achieves the highest percentages of successfully localized images at the ($0.25$m, $2^{\circ}$)
error threshold when considering a budget of $8$k keypoints and it follows the other deep methods closely at the higher error thresholds.
In contrast to R2D2 \cite{r2d2} and ASLFeat \cite{aslfeat}, our network MD-2-Net is trained exclusively using
synthetic homographies and neither on day-night pairs nor on 3D data.

\subsubsection{Image Matching Benchmark \cite{imb}}
This is a recent online benchmark proposed to evaluate the performance of local features \cite{imb}
in the context of \textit{stereo pose recovery} and \textit{multiview reconstruction} on
two sets of sequences, namely Phototourism and PragueParks.
It considers multiple intermediate metrics (Number of Features (NF), Number of Inlier matches (NI),
Repeatability (Rep), Matching Score (MS), Number of inlier Matches filtered by COLMAP \cite{schoenberger2016sfm} (NM),
Number of triangulated Landmarks (NL), Track Length (TL), Absolute Trajectory Error (ATL))
as well as the resulting mean Average Accuracy (mAA) up to $10^{\circ}$.
For a more detailed description of the metrics, we refer to the benchmark documentation \cite{imb}.
We evaluate MD-2-Net on the restricted keypoint category: maximum 2048 keypoints per image.
The benchmark results are reported in Tab.~\ref{tab:imb}.
MD-2-Net achieves competitive results in all the metrics.
In particular, it provides the best mAA on both the sets, for both the stereo and multiview tasks.
A qualitative comparison between the considered methods is provided in Fig.~\ref{fig:imb_teaser} for the stereo pose recovery task.

\subsection{Ablation studies}
\setlength{\textfloatsep}{4pt}
\setlength{\dbltextfloatsep}{4pt}
In order to test the performance of our method with a varying number of detectors,
we train different instances of MD-Net with $N = 1, 2, 4$ and $8$ detectors, using the same primed backbone,
and test them on the HPatches dataset \cite{hpatches}.
It is important to note that $\gamma$, the weight of the dissimilarity loss $\mathcal{L}_\text{dissim}$ plays a crucial role:
the smaller its value, the higher the chances for multiple detectors to find very similar keypoints, and vice versa.
To this purpose, we introduce the \textit{Separability} metric at $n$ pixels, denoted \textit{Sep@n\,\text{px}}.
This measures the overlap between all the detected keypoints as one minus the ratio between the number of keypoints
selected by one detector that are closer than \textit{n} pixels to any other keypoint detected by the other detectors
and the total number of detected keypoints.
The higher the separability, the lower the chances of observing keypoints from different detectors falling withing $n$ pixels
from each other.
As an example, in our test with $N=4$, $\gamma = 2.0$ leads to $\textit{Sep@3\text{px}} = 0.971$
while setting $\gamma = 1.5$ leads to the lower $\textit{Sep@\text{3px}} = 0.81$.
In order to ensure that \textit{Sep@3\text{px}} is higher than $0.95$, we empirically set
$\gamma = 0.5, 2.0$ and $18.0$ for the cases $N = 2, 4$ and $8$, respectively.
The results of our test are reported in Tab.~\ref{tab:hpatches_ablations}
(refer to Sec.~\ref{section:experiments} for more details about the dataset and metrics)
and show that the model trained with two detectors, denoted \textit{MD-2-Net}, offers the best trade-off between the single,
the four and the eight detector versions, in terms of metrics and matching complexity.

When comparing runtimes, matching all the possible pairs between 300 images with 8000 keypoints each
takes 288s with the single detector, 161s using two, 87s using four and only 51s when using eight,
with an average of 6.4ms, 3.6ms, 1.9ms and 1.1ms per pair, respectively.
Tests are carried out on a single Nvidia GTX1080Ti and the matching time considers the scores computation,
the mutual nearest neighbor search and the match aggregation.

Finally, in order to assess the effectiveness of our peakyness loss weighting scheme,
we train our model without the weighting term
$W_{ij}$ in Eq.~\ref{eq:peaky_loss}.
The resulting network, denoted \mbox{\textit{MD-1-Net No-Var}} in Tab.~\ref{tab:hpatches_ablations},
performs considerably worse than \mbox{\textit{MD-1-Net}}, which employs our weighting scheme instead.

\section{Conclusion and future works}
We introduced MD-Net, a novel deep feature extraction network capable of extracting multiple disjointed sets of local features:
these can be matched independently, thus reducing the computational complexity of the matching phase.
The high \textit{separability} values obtained in our analysis,
with varying number of detectors,
confirm the effectiveness
of the novel unsupervised \textit{dissimilarity loss} at the basis of MD-Net.
Additionally, we proposed a variance-based loss dampening scheme that, together with the two-stage training,
avoids the detection of keypoints associated with 
non-discriminative descriptors.

Our experiments show that the network,
trained unsupervised, achieves competitive results on different 3D-related tasks 
at a reduced matching complexity, despite being trained
exclusively on images warped with random homographies.

In the future, we will consider different strategies to select the keypoints at each heatmap, and couple the proposed multi-detector paradigm with a deep matcher architecture,
such as \cite{superglue}, in order to benefit from additional learnt geometric consistency while keeping the
matching cost manageable.
\textbf{Acknowledgement:} This work has been
supported by the FFG, Contract No. 881844: "Pro$^2$Future".

\clearpage
\newpage
\bibliographystyle{IEEEtran}
\bibliography{IEEEabrv,mybib}

\end{document}